\documentclass[letterpaper, 10pt, journal, onecolumn]{IEEEtran}
\usepackage{amsmath,amsfonts,amssymb, amsthm}
\usepackage{algorithmicx}
\usepackage{algpseudocode}
\usepackage{algorithm}
\usepackage{array}
\usepackage{cite}
\usepackage{graphicx}
\usepackage{mathtools}
\usepackage{stfloats}
\usepackage[caption=false,font=normalsize,textfont=sf]{subfig}
\usepackage{textcomp}
\usepackage{url}
\usepackage{verbatim}
\usepackage{xcolor}
\usepackage{booktabs}
\usepackage{float}

\newcommand{\strategy}{SOWL-MPC}

\newcommand{\inv}{^{-1}}
\newcommand{\prev}{\tidx-1}
\renewcommand{\next}{\tidx+1}

\newcommand\oprocendsymbol{\hbox{$\square$}}
\newcommand\oprocend{\relax\ifmmode\else\unskip\hfill\fi\oprocendsymbol}

\newcommand{\RR}{\mathbb{R}}
\newcommand{\NN}{\mathbb{N}}
\newcommand{\EE}{\mathbb{E}}

\newcommand{\cD}{\mathcal{D}}

\newcommand{\cL}{\mathcal{L}}
\newcommand{\cN}{\mathcal{N}}

\newcommand{\cU}{\mathcal{U}}

\newcommand{\cR}{\mathcal{R}}

\newcommand{\postmean}{\mu}
\newcommand{\postcov}{\Sigma}

\newcommand{\priormean}{m}
\newcommand{\kernel}{k}
\newcommand{\gramkernel}{K}

\newcommand{\numdata}{N}

\newcommand{\latentfunc}{\phi}
 
\newcommand{\noise}{w}

\newcommand{\regressor}{x}
\newcommand{\inducingloc}{\bar{x}}
\newcommand{\numinducing}{M}
\newcommand{\inducingmatrix}{\bar{X}}
\newcommand{\inducingval}{\nu}
\newcommand{\designmatrix}{X}
\newcommand{\observationvector}{y}
\newcommand{\dimregressor}{d}
\newcommand{\observation}{y}
\newcommand{\varmean}{m}
\newcommand{\varcov}{S}

\newcommand{\tidx}{t}
\newcommand{\mpcidx}{\tau}

\newcommand{\ptarget}{\hat{\mathrm{y}}}
\newcommand{\pcov}{\Sigma^{\ptarget}}

\newcommand{\radius}{r}

\newcommand{\targetdimstate}{n_z}

\newcommand{\hz}{\hat{\z}}

\newcommand{\x}{\mathrm{x}}
\newcommand{\xt}{\x_t}
\newcommand{\xtp}{\x_{\next}}
\renewcommand{\u}{\mathrm{u}}
\newcommand{\ut}{\u_t}
\newcommand{\edyn}{f}
\newcommand{\dimx}{n_{\x}}
\newcommand{\dimu}{n_{\u}}
\newcommand{\epolicy}{\textsc{mpc}}
\newcommand{\varz}{\sigma_{\z}^2}
\newcommand{\vary}{\sigma_{\v}^2}
\newcommand{\wt}{w_t}

\newcommand{\z}{\mathrm{z}}
\newcommand{\zt}{\z_t}
\newcommand{\tz}{\tilde{\z}}
\newcommand{\tzt}{\tilde{\z}_t}
\renewcommand{\v}{\mathrm{v}}
\newcommand{\vt}{\v_t}
\newcommand{\hv}{\tilde{\v}}
\newcommand{\g}{g}
\newcommand{\h}{H}
\newcommand{\dimz}{n_{\z}}
\newcommand{\dimv}{n_{\v}}
\newcommand{\tpolicy}{\pi}
\newcommand{\htpolicy}{\hat{\pi}}

\newcommand{\zpredt}{\hat{\mathbf{z}}_t}

\newcommand{\Sigmazpredt}{\hat{\boldsymbol{\Sigma}}\vphantom{\boldsymbol{\Sigma}}^\z_t}

\newcommand{\pr}{\theta}
\newcommand{\prt}{\pr_t}

\newcommand{\Xp}{X_+}
\newcommand{\yp}{y_+}
\newcommand{\Sigmap}{\Sigma_+}

\newcommand{\xtt}{\x_{t \mid t}}
\newcommand{\xtpt}{\x_{t+\tau+1 \mid t}}
\newcommand{\xtaut}{\x_{t+\tau \mid t}}
\newcommand{\xTt}{\x_{t+\predhorizon \mid t}}
\newcommand{\utt}{\u_{t \mid t}}
\newcommand{\utaut}{\u_{t+\tau \mid t}}
\newcommand{\uTt}{\u_{t+\predhorizon-1 \mid t}}
\newcommand{\ztp}{\z_{\next}}
\newcommand{\zTt}{\hat{\z}_{t+\predhorizon \mid t}}
\newcommand{\zpredtau}{\hat{\z}_{t+\tau \mid t}}
\newcommand{\predhorizon}{T}
\newcommand{\stagecost}{\ell_{\tau}}
\newcommand{\termcost}{\ell_{\predhorizon}}

\graphicspath{{figs/}}

\begin{document}

\title{
        Safe Learning Predictive Control for Ego-World Robotic Systems
}

\author{
Davide Valenti and Giuseppe Notarstefano%

\thanks{Work partially funded by the European Union - Next Generation EU - under the National Recovery and Resilience Plan (NRRP), Mission 4, Component 2, Investment 3.3; CUP J33C24001490009 and by SACMI S.C.}
\thanks{
Authors are with the Department of Electrical, Electronic and
Information Engineering, University of Bologna,
Bologna, Italy, 
        {\tt\small \{d.valenti, giuseppe.notarstefano\}@unibo.it}.
}
}

\maketitle

\theoremstyle{definition}
\newtheorem{assumption}{Assumption}
\newtheorem{problem}{Problem}
\newtheorem{remark}{Remark}
\newtheorem{lemma}{Lemma}
\newtheorem{property}{Property}
\newtheorem{modeling_assumption}{MA}

\begin{abstract}
  Safe autonomous navigation in shared environments requires the
  ability to anticipate and react to the latent behaviors of
  surrounding robots.
  In this paper, we propose \strategy{}, a safe learning-based
  predictive control strategy for a novel scenario, which we name
  \textsc{ego--world} robotic framework. In this setting, the control policy of
  the world robot is unknown and the ego exploits data to learn it and
  perform safe maneuvers.
  The proposed architecture combines an online learning mechanism
  based on Sparse Variational Gaussian Processes (SVGPs) with a
  receding-horizon control scheme.
  Relying solely on noisy state measurements, our approach infers a
  posterior distribution over the latent world policy, which is
  updated on streaming data via Online Variational Conditioning (OVC).
  The learned policy is propagated through the nonlinear world
  dynamics using an approximate moment propagation scheme, and fed to
  an uncertainty-aware Model Predictive Control (MPC), thus enabling safe
  maneuvering of the ego robot.
  The real-time feasibility and safety guarantees of \strategy{} are
  demonstrated through extensive Monte Carlo virtual experiments in ROS 2, and
  validated on real-world robotic hardware in an indoor arena.
  \\
  \\
  Video: \url{https://youtu.be/nopsMYRULWs}
  
\end{abstract}


\section{Introduction}
From high-speed autonomous racing to
efficient logistics in smart warehouses, modern robotic systems are
increasingly required to operate in highly dynamic and uncertain
environments~\cite{hewing2020learning,brunke2022safe}.
To address these challenges, a dyadic modeling setting can be introduced,
consisting of a controlled \textit{ego} robot and an uncooperative,
partially unknown \textit{world} robot.
We refer to this novel framework as \textsc{ego--world} paradigm, in
which safe maneuvering of the ego robot demands the ability to
anticipate and react to the latent behavior of the world robot.
While existing approaches rely on fixed motion models or offline-trained predictors, such assumptions can lead to overly conservative or unsafe behavior when robots deviate from expected dynamics.
To overcome these limitations, we propose a safe learning-based predictive control strategy that infers the latent world robot policy directly from online observations and continuously adapts as new data are collected.
\begin{figure}[thpb]
        \centering
        \includegraphics[scale=1.0]{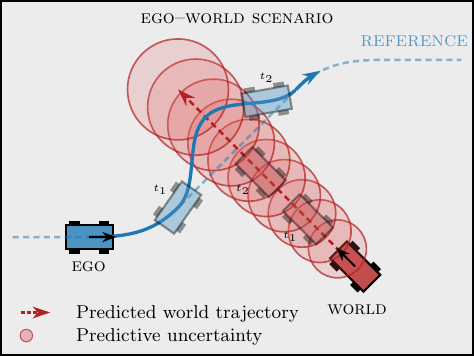}
        \caption{Overview of the considered \textsc{ego--world} scenario. The ego
            vehicle (blue) exploits the predicted trajectory of the
            world vehicle (red) and its associated predictive
            uncertainty (shaded red) to perform collision-free
            maneuvers.}
        \label{fig:scenario}
\end{figure}
\paragraph*{Related Work}
Traditionally, collision avoidance in presence of partially unknown robots has been addressed through reactive strategies, which rely on instantaneous sensing rather than motion forecasting. 
Foundational approaches include the Dynamic Window Approach~\cite{fox2002dynamic}, which selects admissible velocity commands over a short horizon, and Artificial Potential Fields~\cite{khatib1986real}, recently augmented with optimization-based filters to handle fast-moving obstacles~\cite{liu2022real}.
Similarly, methods operating in the velocity space, such as Optimal Reciprocal Collision Avoidance~\cite{van2008reciprocal} and modulation-based approaches~\cite{huber2022fast}, compute a set of safe velocities from the observed states of surrounding robots. 
Reactive safety guarantees have also been formalized using Control Barrier Functions (CBFs)~\cite{ames2019control}, where the control input is minimally corrected to ensure the forward invariance of a safe set.
Recent works have combined CBFs with learning techniques, including Gaussian Processes (GPs)~\cite{keyumarsi2023lidar} and Neural Networks~\cite{harms2024neural}, to synthesize barrier functions directly from raw sensor data. 
In parallel, Deep Reinforcement Learning~\cite{tai2017virtual} has been explored to learn end-to-end policies mapping sensor measurements directly to safe navigation commands.
Despite their computational efficiency, reactive methods are inherently myopic, and do not exploit the dynamic behavior of surrounding robots in a predictive way.
Early approaches to extend reactive strategies focused exclusively on motion forecasting, providing open-loop predictions of surrounding robots without integrating them into predictive control frameworks.
Representative examples include~\cite{lefkopoulos2020interaction}, where an Interacting Multiple Model Kalman Filter is used for vehicle motion prediction in highway driving scenarios, and~\cite{mascaro2024robot}, proposing a Transformer-based architecture for human motion forecasting.
Building upon these ideas, subsequent works coupled motion forecasting with planning algorithms to generate collision-free
trajectories.
Within this paradigm, a first line of work concerns interaction-aware methods, which explicitly account for the influence of the ego robot on the future behavior of surrounding agents.
For instance, neural network-based predictions have been integrated
into decentralized Model Predictive Control (MPC) schemes for multi-robot collision avoidance~\cite{zhu2021learning,
  gupta2023interaction}.
However, these approaches rely on predictors trained offline, preventing adaptation to previously unseen behaviors during deployment.
Alternatively,~\cite{streichenberg2023multi} uses Model Predictive Path Integral (MPPI) to sample candidate control sequences for autonomous vessels, and discards those leading to collisions, thereby avoiding the need for pre-trained prediction models.
Nevertheless, the sampling strategy is not updated online from streaming observations, and assumes all robots behave rationally according to the same underlying cost function.
The same rationality assumption underpins game-theoretic methods, which formally model multi-robot interactions by recasting the motion planning problem as a dynamic game.
The trajectories of all robots are then jointly optimized by computing an approximate Nash equilibrium of the game, e.g., in~\cite{fisac2019hierarchical} to inform a lower-level planner, or in~\cite{kalaria2025real} for infinite-horizon racing formulations.
These methods, however, do not exploit data to learn the behavior of the interacting robots online, and remain largely computationally intractable in real-world settings.
Finally, probabilistic planning methods account for uncertainty in the predicted motion of other robots.
Relevant works include chance-constrained MPC formulations that limit collision probabilities with dynamic obstacles~\cite{zhu2019chance}, and scenario-based predictive control schemes leveraging probabilistic high-order Markov chains~\cite{oleinikov2024scenario}.
While these approaches propagate predictive uncertainty, they assume that the underlying prediction model is known a priori and do not dynamically update it using observed data.
Gaussian Processes (GP)~\cite{williams2006gaussian} are an important
building block for our proposed strategy, but in the robotic
literature they have been exclusively used to learn unmodeled dynamics
of the controlled system itself.
Within this single system setting, GPs have been successfully
integrated into predictive control frameworks, as surveyed
comprehensively in~\cite{scampicchio2025gaussian}.
A recurring pattern in the literature involves decoupling the dynamics of an unknown system into a nominal component, e.g., derived from first principles, and a residual term, learned via GP regression.
This paradigm has been integrated into finite-horizon optimal control~\cite{sforni2021learning} and Model Predictive Control schemes, employing either reachable sets~\cite{hewing2019cautious} or sampling-based evaluations within a Sequential Quadratic Programming (SQP) framework~\cite{prajapat2024towards}.
More recently, \cite{bartels2026real} extended this GP-based MPC paradigm with spatio-temporal approximations, enabling online learning on streaming data at fixed computational cost.
All the above, however, focus on learning the dynamics of the controlled system itself, rather than forecasting the motion of other robots in the environment.
Closer to our \textsc{ego--world} setting,~\cite{brudigam2021gaussian, yoon2021interaction} integrate GP-based predictions of the closed-loop dynamics of an exogenous vehicle within Stochastic MPC formulations.
Building upon this setting,~\cite{zhu2023gaussian} trains a GP model on prior racing data to model the closed-loop interaction dynamics between two autonomous vehicles, where the control policy of the exogenous vehicle is unknown.
Despite these advances, such techniques rely solely on offline
training and assume specific interaction patterns
(e.g., lane changing), thus limiting their applicability to previously
observed behaviors.
\paragraph*{Contributions}
The main contributions of this paper are threefold. 
First, we propose Safe Online World policy Learning Model Predictive Control (\strategy{}), a novel learning-based safe control strategy for \textsc{ego--world} robotic systems. 
In the proposed framework, the ego robot is controlled via a receding-horizon scheme relying on a predictive model of the world robot, which exploits prior knowledge of its nominal dynamics and is refined online on observed data.
Rather than identifying open- or closed-loop dynamics, our strategy focuses on learning the latent state-feedback policy governing the world robot behavior.
This formulation leverages the structural decoupling between the world robot dynamics, typically time-invariant and reasonably approximated, and the control law, unknown and potentially non-stationary. 
Furthermore, since the input space is generally lower-dimensional than the state space, inferring the control policy results in a more data-efficient learning problem.
Second, by modeling the unknown world policy using Sparse Variational Gaussian Processes (SVGPs), we develop an online update scheme based on Online Variational Conditioning (OVC), recently proposed in~\cite{maddox2021conditioning} for statistical inference problems.
The resulting strategy is combined with an approximate moment propagation scheme to obtain a multi-step probabilistic forecast of the future world robot trajectory, along with a measure of its predictive uncertainty.
To the best of our knowledge, this is the first application of OVC to online policy identification and trajectory prediction in robotic systems.
Third, we demonstrate the real-time applicability of the proposed framework by embedding it into a full-stack ROS 2 implementation, seamlessly integrated within \textsc{ChoiRbot}~\cite{testa2021choirbot}, a toolbox for distributed robotics.
We perform realistic virtual experiments involving two NVIDIA JetRacer autonomous cars, and conduct extensive Monte Carlo numerical tests to study the performance of our strategy in terms of prediction accuracy and safety guarantees.
Finally, \strategy{} is deployed on real hardware and evaluated in a challenging scenario involving multiple overtaking and crossing maneuvers.

\paragraph*{Organization}
The paper unfolds as follows. 
Section~\ref{sect:problem_formulation} introduces the problem of safe control for \textsc{ego--world} robotic systems and provides the necessary preliminaries on Gaussian Process Regression. 
Our learning strategy is then presented in Section~\ref{sect:solution_strategy}, along with its integration within a receding-horizon control framework. 
Finally, virtual and real experiments are reported in Sections~\ref{sect:virtual_experiments} and ~\ref{sect:real_experiments}, respectively.

\paragraph*{Notation}
$I_n$ is the $n \times n$ identity matrix, while $\RR_+$ is the set of non-negative real numbers.
Given $Q = Q^\top \succeq~0, Q \in \RR^{n \times n}$, and $x \in \RR^n$, we denote the $Q$-norm of $x$ as $\|x\|_Q = \sqrt{x^\top Q x}$.
The symbol $\text{col}(v_1, \ldots, v_N)$ denotes the concatenation of the vectors $v_1, \ldots, v_N$.
Given $v_1, \ldots, v_N \in \RR$, we denote as $\text{diag}(v_1, \ldots, v_N)$ the diagonal matrix with $v_i$ as its $i$-th diagonal component.
$\cN(\mu, \Sigma)$ denotes a multivariate Gaussian distribution with mean $\mu \in \RR^n$ and covariance matrix $\Sigma \in \RR^{n \times n}$. 
Given two matrices $X = [x_1, \ldots, x_N]^\top \in \RR^{N \times n}$ and $X^\prime = [x^\prime_1, \ldots, x^\prime_{N^\prime}]^\top \in \RR^{N^\prime \times n}$, we denote by $K_{XX^\prime} \in \RR^{N \times N^\prime}$ the Gram matrix associated with the kernel function $k: \RR^n \times \RR^n \to \RR$, whose $(i,j)$-th entry is $k(x_i, x^\prime_j)$.
\section{Problem Statement \& Preliminaries}
\label{sect:problem_formulation}
\subsection{Safe Control of Ego--World Robotic Systems} 
In this paper, we consider the problem of safe control for an \textit{ego} robot, navigating in a shared environment with a \textit{world} robot, whose control policy is unknown.
A graphical illustration of the considered scenario is depicted in Fig.~\ref{fig:scenario}.
Formally, the ego robot is described as a discrete-time nonlinear system:
\begin{equation}
        \label{eq:ego_system}
        \xtp = \edyn(\xt, \ut),
\end{equation}
where $\xt \in \RR^{\dimx}$ and $\ut \in \cU \subseteq \RR^{\dimu}$ are the ego states and inputs at time $\tidx \in \NN$, respectively, and $f: \RR^{\dimx} \times \cU \to \RR^{\dimx}$ describes its dynamics.
As for the world robot, it evolves according to an input-affine nonlinear dynamics:
\begin{equation}
        \label{eq:world_system}
        \ztp = \g(\zt) + \h(\zt) \vt.
\end{equation} 
Here, $\zt \in \RR^{\dimz}$ and $\vt \in \RR^{\dimv}$ are states and inputs of the world system at time $\tidx \in \NN$, respectively, while $\g(\zt): \RR^{\dimz} \to \RR^{\dimz}$ and $\h(\zt): \RR^{\dimz} \to \RR^{\dimz \times \dimv}$ denote its drift vector and control matrix, which we assume to be continuously differentiable.
To introduce the problem scenario of interest for \textsc{ego--world} robotic systems~\eqref{eq:ego_system}-\eqref{eq:world_system}, we assume the following.
\vspace{1ex}
\\
{\bf Modeling Assumptions (MA)} 
\begin{modeling_assumption}[World system policy]
        \label{ass:state_feedback_policy}
        The world system evolves according to a state-feedback policy $\tpolicy(\zt):\RR^{\dimz}\to\RR^{\dimv}$. 
        \oprocend
\end{modeling_assumption}
\begin{modeling_assumption}[Ego-available information]
        \label{ass:measurement_model}
        The ego system: (i) has prior knowledge of $\g(\zt)$ and $\h(\zt)$, (ii) it is entirely agnostic to the world system control policy $\tpolicy(\zt)$, and (iii) cannot directly measure $\vt$, but only noisy observations
        \begin{equation}
                \label{eq:observation_model}
                \tzt = \zt + \wt, \quad \wt \sim \cN(0, \varz I_{\dimz})
        \end{equation} 
        of the world system state, corrupted by white noise with variance $\varz \in \RR_+$.
        \oprocend
\end{modeling_assumption}

\noindent In this paper, we focus on designing a receding-horizon control law for the ego system, in order to guarantee safe maneuvering in presence of the world robot.
Specifically, let us denote as $\mathbf{z}_t \coloneq \text{col}(\z_{t+1}, \ldots, \z_{t+\predhorizon})$ the true future trajectory of the world system, starting from time $t$, over the prediction horizon $\predhorizon \in \NN$.
If $\mathbf{z}_t$ were available, safe control of the ego system could be achieved by a model predictive control scheme, obtained by solving at each time $t$ the following optimal control problem:
\begin{subequations}\label{eq:ego_controller}
\begin{align}
        \min_{\substack{\xtt, \ldots, \xTt \\ \utt, \ldots, \uTt}} &\sum_{\mpcidx = 0}^{\predhorizon - 1} \stagecost(\xtaut, \utaut) + \termcost(\xTt) \label{eq:cost}\\
        \text{s.t.} \qquad \qquad & \hspace{-7ex}\xtpt = \edyn(\xtaut, \utaut), \: \mpcidx=0,\ldots,\predhorizon\!-\!1, \label{eq:dynamics_constraints}\\
        & \hspace{-7ex}\xtt = \xt, \label{eq:initcond_constraints}\\
        & \hspace{-7ex}\utaut \in \cU, \qquad \qquad \qquad \quad \: \: \mpcidx = 0, \ldots, \predhorizon\!-\!1, \label{eq:input_constraints} \\
        & \hspace{-7ex}\text{dist}(\xtaut, \z_{t+\tau}) \geq 0, \qquad \quad \mpcidx = 1, \ldots, \predhorizon,\label{eq:ca_constraints}
\end{align}
\end{subequations}
where $\stagecost: \RR^{\dimx} \times \cU \to \RR$, $\termcost: \RR^{\dimx} \to \RR$ are twice differentiable stage and terminal cost functions.
The first element $\ut = \utt^\star$ of the optimal input trajectory, obtained by solving problem~\eqref{eq:ego_controller}, is then applied to the ego system, and the procedure is repeated at the next time step in a receding-horizon fashion.
However, collision avoidance constraints~\eqref{eq:ca_constraints}, whose explicit expression will be detailed in Section~\ref{subsect:mpc}, depend on the future trajectory of the world system, which is not available under Modeling Assumption MA~\ref{ass:measurement_model}.
As a result, the ideal control law~\eqref{eq:ego_controller} cannot be implemented.

\noindent To overcome this limitation, we equip the ego system with a proxy $\htpolicy_{\prt}(\zt)$ of the true policy $\tpolicy(\zt)$, which is used to generate predictions of the closed-loop world trajectory $\zpredtau \in \RR^{\dimz}$ with associated predictive uncertainty $\hat{\Sigma}^{\z}_{t+\tau\mid t} \in \RR^{\dimz \times \dimz}$, motivating the following problem statement.
\begin{problem}[\small\textsc{Safe maneuvering in ego--world systems}]
        \label{prob:online_learning_safe}
        Consider the \textsc{ego--world} system given by~\eqref{eq:ego_system}-\eqref{eq:world_system} and characterized by Modeling Assumptions MA~\ref{ass:state_feedback_policy} and MA~\ref{ass:measurement_model}. Our objective is to design an online framework to:
        \begin{enumerate}
                \item[a)] Iteratively update the parameters $\prt$ of a proxy $\htpolicy_{\prt}$ of the true policy $\tpolicy$, based on noisy observations~\eqref{eq:observation_model}.
                \item[b)] Generate, at each time step $t$, a finite-horizon prediction $\zpredt \coloneq \text{col}(\hat{\z}_{t+1\mid t}, \ldots, \zTt)$ of the world system trajectory with associated predictive uncertainty $\Sigmazpredt \coloneq \{\hat{\Sigma}^\z_{t+\tau \mid t}\}_{\tau=1}^{\predhorizon}$ leveraging the estimated policy $\htpolicy_{\prt}$.
                \item[c)] Implement an uncertainty-aware model predictive control law $\ut = \epolicy(\xt, \zpredt, \Sigmazpredt)$, enabling tracking of a reference trajectory while ensuring safe maneuvering.
        \end{enumerate}
\end{problem}
\subsection{Gaussian Processes for Regression}
We start by briefly recalling the fundamentals of non-parametric regression via Gaussian Processes (GP) \cite{williams2006gaussian}.
Consider a real-valued function $\latentfunc(x): \RR^\dimregressor \to \RR$, and assume $\numdata$ samples $X = [x_1, \ldots, x_{\numdata}]^\top \in \RR^{\numdata \times \dimregressor}, y = [y_1, \ldots, y_{\numdata}]^\top \in \RR^{\numdata}$ have been collected according to the observation model $\observation_i = \latentfunc(\regressor_i) + \noise_i, \noise_i \sim \cN(0, \sigma^2_y)$, with noise variance $\sigma_y^2 \in \RR_+$.
The core principle of GP regression is to model the latent function $\latentfunc$ as a Gaussian Process, denoted as:
\begin{equation}
        \label{eq:gp_assumption}
        \latentfunc(\regressor) \sim \mathcal{GP}(\priormean(\regressor), \kernel(\regressor, \regressor')),
\end{equation}
where $\priormean: \RR^\dimregressor \to \RR$ and $\kernel: \RR^\dimregressor \times \RR^\dimregressor \to \RR$ are the prior mean and kernel functions, respectively.
In this paper, we consider a constant prior mean $m(x) = \bar{m}$ and a Radial Basis Function (RBF) kernel $k(x,x^\prime) = \sigma_\phi^2 \exp\left(-1/2 \|x - x'\|_{\Lambda^{-1}}^2\right)$, and treat $\eta \coloneq \{\bar{m}, \sigma_\phi^2, \Lambda, \sigma_y^2\}$ as hyperparameters to be selected.
Under the GP assumption~\eqref{eq:gp_assumption}, given a set of unseen test inputs $\designmatrix^\star = [x^\star_1, \ldots, x^\star_{N^\star}]^\top \in \RR^{\numdata^\star \times \dimregressor}$, the posterior predictive distribution $p(\latentfunc(\designmatrix^\star) \mid \observationvector)$ is Gaussian, with mean and covariance given by:
\begin{subequations}\label{eq:gp_posterior}
\begin{align}
        \postmean^\star&=\priormean(\designmatrix^\star)+\gramkernel_{\designmatrix\designmatrix^\star}^\top(\gramkernel_{\designmatrix\designmatrix}+\sigma_y^2 I_\numdata)\inv (\observationvector - \priormean(\designmatrix)), \label{eq:gp_posterior_mean}\\
        \postcov^\star&=\gramkernel_{\designmatrix^\star\designmatrix^\star}-\gramkernel_{\designmatrix\designmatrix^\star}^\top(\gramkernel_{\designmatrix\designmatrix}+\sigma_y^2 I_\numdata)\inv \gramkernel_{\designmatrix\designmatrix^\star}, \label{eq:gp_posterior_covariance}
\end{align}
\end{subequations}
where $\latentfunc(\designmatrix^\star) = [\latentfunc(x^\star_1), \ldots, \latentfunc(x^\star_{N^\star})]^\top \in \RR^{\numdata^\star}$ and $\priormean(\designmatrix^\star) = [\priormean(x^\star_1), \ldots, \priormean(x^\star_{\numdata^\star})]^\top \in \RR^{\numdata^\star}$.
The major drawback of GP regression methods is their prohibitive computational cost, typically $O(N^3)$ for training and $O(N^2)$ for prediction~\cite{williams2006gaussian}, making this formulation impractical when dealing with large datasets or streaming data.
To reduce the computational complexity, inducing-point methods~\cite{snelson2005sparse} have emerged as a popular approach for approximate inference.
The key idea is to introduce $\numinducing \ll \numdata$ \textit{inducing locations} $\inducingmatrix = [\inducingloc_1, \ldots, \inducingloc_\numinducing]^\top \in \RR^{\numinducing \times \dimregressor}$, with corresponding \textit{inducing values} $\inducingval = [\latentfunc(\inducingloc_1), \ldots, \latentfunc(\inducingloc_\numinducing)]^\top \in \RR^\numinducing$.
Under the conditional independence assumption, i.e. $\latentfunc(X) \perp \latentfunc(X^\star) \mid \inducingval$, the posterior predictive distribution can be computed by marginalizing over $\inducingval$, yielding:
\begin{equation}
        p(\latentfunc(X^\star) \mid \observationvector) = \int p(\latentfunc(X^\star) \mid \inducingval) p(\inducingval \mid \observationvector) d\inducingval. \label{eq:posterior_predictive_inducing}
\end{equation}
The expression in eq.~\eqref{eq:posterior_predictive_inducing} requires evaluating the posterior over inducing values $p(\inducingval \mid \observationvector)$, which is still Gaussian with mean and covariance matrix given by:
\begin{subequations}\label{eq:posterior_inducing}
\begin{align}
        m^\star &= \priormean(\inducingmatrix) + \gramkernel_{X\inducingmatrix}^\top \left(\gramkernel_{XX}+\sigma_y^2 I_N\right)\inv (y - \priormean(X)) \label{eq:posterior_inducing_mean}\\
        S^\star &= \gramkernel_{\inducingmatrix\inducingmatrix}-\gramkernel_{X\inducingmatrix}^\top \left(\gramkernel_{XX}+\sigma_y^2 I_N\right)\inv \gramkernel_{X\inducingmatrix}\label{eq:posterior_inducing_cov}. 
\end{align}
\end{subequations}
However, since the exact evaluation of~\eqref{eq:posterior_inducing} features cubic complexity, Sparse Variational Gaussian Processes (SVGPs)~\cite{titsias2009variational} approximate $p(\inducingval \mid \observationvector)$ with a variational distribution $q(\inducingval)$.
Since $p(\inducingval \mid \observationvector) \sim \cN(m^\star, S^\star)$, $q(\inducingval)$ is posited as a Gaussian distribution with mean $\varmean \in \RR^\numinducing$ and covariance matrix $\varcov \in \RR^{\numinducing \times \numinducing}$.
The variational moments $\varmean, \varcov$, along with the inducing locations $\inducingmatrix$, are then learned from data by maximizing the evidence lower bound (ELBO) loss function~\cite{hensman2015scalable}, as detailed in Section~\ref{sect:solution_strategy}.
Taking the integral in eq.~\eqref{eq:posterior_predictive_inducing} in light of the variational approximation, we obtain an approximate posterior predictive distribution $q(\phi(X^\star)) \sim \cN(\postmean, \postcov)$ with mean and covariance given by:
\begin{subequations}\label{eq:svgp_posterior}
\begin{align}
        \postmean&=\priormean(\designmatrix^\star) + \gramkernel_{\inducingmatrix\designmatrix^\star}^\top \gramkernel_{\inducingmatrix\inducingmatrix}\inv (\varmean - \priormean(\inducingmatrix)), \label{eq:svgp_posterior_mean} \\
        \postcov &= \gramkernel_{\designmatrix^\star \designmatrix^\star}\!-\!\gramkernel_{\inducingmatrix\designmatrix^\star}^\top \gramkernel_{\inducingmatrix\inducingmatrix}\inv \left(\gramkernel_{\inducingmatrix\inducingmatrix} - \varcov \right) \gramkernel_{\inducingmatrix\inducingmatrix}\inv\gramkernel_{\inducingmatrix\designmatrix^\star}. \label{eq:svgp_posterior_covariance}
\end{align}
\end{subequations}
\begin{remark}
        Approximate inference via inducing-point methods reduces the computational cost to $O(NM^2)$ and $O(M^2)$ for training and prediction, respectively \cite{snelson2005sparse}. \oprocend
\end{remark}
\section{Solution Strategy}
\label{sect:solution_strategy}
We propose a solution to Problem~\ref{prob:online_learning_safe}, introduced in Section~\ref{sect:problem_formulation}, based on modeling each component $\pi^i(\z): \RR^{\dimz} \to \RR, \, i \in \{1, \ldots, \dimv\}$ of the unknown world policy $\tpolicy$ as a separate, independent SVGP model. 
To maintain notational consistency throughout the rest of the paper, we denote as $\htpolicy^i_{\pr^i_t}(\z)\sim \cN(\postmean^{i}_{\pr^i_t}(\z), \postcov^{i}_{\pr^i_t}(\z))$ the univariate posterior distribution~\eqref{eq:svgp_posterior} of the $i$-th SVGP evaluated at $\z$ with parameters $\prt^i \coloneq \{\eta^i_t, \inducingmatrix^i_t, \varmean^i_t, \varcov^i_t\}$ (cf. Section~\ref{sect:problem_formulation}).
Consequently, denoting the stack of all parameters at time $t$ as $\prt = \{\prt^i\}_{i = 1}^{\dimv}$, the proxy of the world policy evaluated at $\z$ is a multivariate Gaussian $\htpolicy_{\prt}(\z) \sim \cN(\postmean_{\prt}(\z), \postcov_{\prt}(\z))$, with posterior mean $\postmean_{\prt}(\z): \RR^{\dimz} \to \RR^{\dimv}$ and covariance $\postcov_{\prt}(\z): \RR^{\dimz} \to \RR^{\dimv \times \dimv}$ defined as:
\begin{subequations}
\begin{align}
    \postmean_{\prt}(\z) &= [\postmean^1_{\prt^1}(\z), \ldots, \postmean^{\dimv}_{\prt^{\dimv}}(\z)]^\top, \\
    \postcov_{\prt}(\z) &= \text{diag}(\postcov^1_{\prt^1}(\z), \ldots, \postcov^{\dimv}_{\prt^{\dimv}}(\z)).
\end{align}
\end{subequations}
Our strategy follows a two-stage approach, consisting in an offline and an online phase.
\emph{Offline}, we select the initial hyperparameters $\pr_0$ of the SVGP models by maximizing a lower bound on the marginal likelihood over a previously collected pretraining dataset $\cD$. 
\emph{Online}, SVGP parameters $\prt$ are recursively updated via Online Variational Conditioning (OVC) on streaming data, to improve the prediction accuracy of the proxy policy $\htpolicy_{\prt}$.
Specifically, at each time step $t$, the ego robot performs (i) an input reconstruction step, to generate an estimate $\hv_{\prev} \in \RR^{\dimv}$ of $\v_{\prev}$ from consecutive noisy state observations $(\tz_{\prev}, \tzt)$, (ii) a conditioning procedure, yielding updated $\prt$ and (iii) a prediction rollout, by forward integration of the world system dynamics and moment propagation.
The resulting finite-horizon prediction $\zpredt$, with predictive uncertainty $\Sigmazpredt$, is then incorporated into a nonlinear MPC framework that enables the ego system to execute safe maneuvers.
The \strategy{} procedure is summarized in Algorithm~\ref{algo:nome_strategia} and detailed in forthcoming subsections. 

\begin{algorithm}
\caption{\strategy \ -- (Ego robot)}
\label{algo:nome_strategia}
\begin{algorithmic}[htpb]
    \State Initialization: $\x_0, \tz_0 \in \RR^{\dimx} \times \RR^{\dimz}$, pretraining dataset $\cD$
    \State \textbf{Offline Pretraining:} $\pr_0 = \arg\min_\pr \cL(\pr)$ 

    \For{$t = 1, 2, ...$}
        \State \textbf{Measurement and Input Reconstruction}
        \State Observe $\tz_t = \zt + \wt$
        \State $\hv_{\prev} = \h(\tz_{\prev})^\dagger (\tzt - \g(\tz_{\prev}))$
        \\
        \State \textbf{Online Conditioning and Trajectory Rollout}
        \State $\prt = $ \textsc{ovc}$\left(\pr_{\prev}, \tz_{\prev}, \hv_{\prev}\right)$ \Comment{Algorithm~\ref{algo:ovc}}
        \State $\zpredt, \Sigmazpredt = $ \textsc{rollout}$\left(\tz_t, \htpolicy_{\prt} \right)$ \Comment{Algorithm~\ref{algo:rollout}}
        \\
        \State \textbf{Safe Predictive Controller}
        \State $\ut = \epolicy(\xt, \zpredt, \Sigmazpredt)$
        \State $\x_{\next} = f(\xt, \ut)$ 
    \EndFor
\end{algorithmic}
\end{algorithm}

\begin{figure}[htpb]
  \centering
  \includegraphics[width=0.5\textwidth]{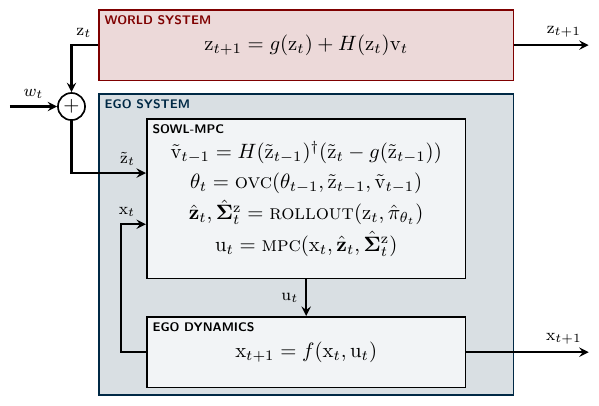}
  \caption{Block diagram of the \strategy{} architecture, illustrating the closed-loop information flow. At each time step $t$, the ego system acquires a noisy measurement $\tzt$ of the world system, updates the parameters $\prt$ of the estimated policy $\htpolicy_{\prt}$ and performs a multi-step rollout of the world system trajectory. The resulting $\zpredt$ and its associated uncertainty $\Sigmazpredt$ are integrated in a nonlinear MPC scheme to compute a safe input $\ut$ for the ego system.}
  \label{fig:blockscheme}
\end{figure}

\subsection{Offline Pretraining}
\label{subsect:offline_pretraining}
This step provides an informed prior for the subsequent online updates by selecting the initial kernel and variational hyperparameters $\pr^i_0, i \in \{1, \ldots, \dimv\}$ of the SVGP models.
Specifically, we consider a pretraining dataset $\cD = \{\mathrm{Z}, \mathrm{y}\}$ of $N$ noisy measurements of the world system, where $\mathrm{Z} \coloneq [\tz_1, \ldots, \tz_\numdata]^\top \in \RR^{\numdata \times \dimz}$ are collected according to the observation model~\eqref{eq:observation_model}.
The corresponding (noisy) inputs $\mathrm{y} \coloneq [\hv_1, \ldots, \hv_\numdata]^\top \in \RR^{\numdata \times \dimv}$ for pretraining can be reconstructed from 
$\mathrm{Z}$ by leveraging knowledge of the world system dynamics (cf. Modeling Assumption MA~\ref{ass:measurement_model}), e.g. via a least-squares approach, as discussed in Section~\ref{subsect:input_reconstruction}.
Since each component of the unknown policy is modeled as an independent SVGP, the pretraining problem features a separable structure:
\begin{equation}
    \label{eq:pretraining_elbo}
    \pr_0 = \arg\min_{\pr} \cL(\pr) = - \sum_{i = 1}^{\dimv} \cL^i (\pr^i).
\end{equation}
Here, $\cL^i(\pr^i)$ is the Evidence Lower Bound (ELBO)~\cite{hensman2015scalable} for the $i$-th SVGP which, by denoting as $\mathrm{y}^i \in \RR^{\numdata}, i \in \{1, \ldots, \dimv\}$ the columns of $\mathrm{y}$, is defined as:
\begin{equation*}
    \cL^i(\pr^i)= \EE_{\htpolicy^i_{\pr^i}(\mathrm{Z})}\left[ \log p(\mathrm{y}^i\mid\tpolicy^i(\mathrm{Z})) \right] - \operatorname{KL}(q(\inducingval^i)\,\|\,p(\inducingval^i)),
\end{equation*}
where $\tpolicy^i(\mathrm{Z}) \coloneq [\tpolicy^i(\tz_1), \ldots, \tpolicy^i(\tz_\numdata)]^\top \in \RR^\numdata$ and $\text{KL}$ denotes the Kullback-Leibler divergence between probability distributions.
Similarly to the exact log-marginal likelihood for exact GPs, the ELBO represents a trade-off between data fit and model complexity.
In detail, the expected log-likelihood term $\EE_{\htpolicy^i_{\pr^i}(\mathrm{Z})}\left[ \log p(\mathrm{y}^i\mid\tpolicy^i(\mathrm{Z})) \right]$ measures the accuracy of the model with respect to the observed data under the approximate variational posterior $\htpolicy^i_{\pr^i}$. 
On the other hand, $\operatorname{KL}(q(\inducingval^i) \,\|\, p(\inducingval^i))$ intuitively acts as a regularizer by penalizing deviations of the variational distribution $q(\inducingval^i) \sim \cN(\varmean^i, \varcov^i)$ from the prior over inducing values $p(\inducingval^i)$, thereby reducing the risk of overfitting. 
In this paper, we keep the kernel hyperparameters $\eta^i$ fixed to the pretraining optimal values $\eta^i_0$, computed by solving problem~\eqref{eq:pretraining_elbo}.
On the other hand, the inducing locations $\inducingmatrix^i_t$ and variational moments $\varmean^i_t, \varcov^i_t$ of each SVGP are updated online as new measurements are acquired.
\subsection{Measurement and Input Reconstruction}
\label{subsect:input_reconstruction}
\begin{figure}[htbp]
    \centering
    \includegraphics[scale=1.0]{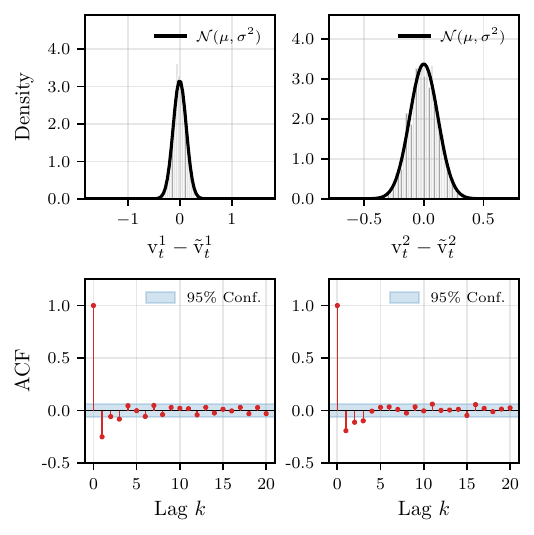}
    \caption{Whiteness test for the input reconstruction error $\vt - \hv_t$ obtained by solving problem~\eqref{eq:input_reconstruction} with $N = 1000$ noisy state measurements $\tzt$ with standard deviation $\sigma_\z = 0.01$. Inputs are longitudinal ($\vt^1$) and angular ($\vt^2$) velocities of a differential drive world system. Top row: empirical distribution of the reconstruction error for each input dimension and Gaussian fit. Bottom row: autocorrelation function of the reconstruction error.}
    \label{fig:input_statistics}
\end{figure}

During online operation, to overcome the lack of direct input observations (cf. Modeling Assumption MA~\ref{ass:measurement_model}), the ego robot needs to reconstruct an estimate $\hv_{\prev}$ of the world system input by relying solely on consecutive measurements $(\tz_{\prev}, \tzt)$, obtained according to~\eqref{eq:observation_model}.
We perform this reconstruction by solving online the following one-step least-squares problem:
\begin{equation}
    \label{eq:input_reconstruction}
    \hv_{\prev} = \arg\min_{\v} \|\tzt - \g(\tz_{\prev}) - \h(\tz_{\prev})\v\|^2.
\end{equation}
Indeed, assuming $\h(\tz_{\prev})$ is tall ($\dimv < \dimz$) and full-column rank, the solution of problem~\eqref{eq:input_reconstruction} exists and is unique, and follows from stationarity conditions:
\begin{equation}
    \label{eq:solution_input_reconstruction}
    \hv_{\prev} = \h(\tz_{\prev})^\dagger (\tzt - \g(\tz_{\prev})),
\end{equation}
where $\h(\tz_{\prev})^\dagger \coloneq (\h(\tz_{\prev})^\top \h(\tz_{\prev}))^{-1} \h(\tz_{\prev})^\top$ denotes the left pseudoinverse of $\h(\tz_{\prev})$.
We remark that, since the world system dynamics~\eqref{eq:world_system} is a nonlinear function of the state, the solution to problem~\eqref{eq:input_reconstruction} is not, in general, Gaussian distributed.
Nevertheless, we approximate the distribution of reconstructed inputs~\eqref{eq:solution_input_reconstruction} as a multivariate Gaussian in order to apply Online Variational Conditioning for the subsequent online updates.
Indeed, under moderate noise, the nonlinear dynamics~\eqref{eq:world_system} of the world system is typically well described by its local linearization around $\tz_{\prev}$, making the approximation reasonable, as shown in Fig.~\ref{fig:input_statistics} for a differential drive world system.
Specifically, we observe that the empirical distribution of the reconstruction error $\hv_t - \vt$ closely matches a zero-mean Gaussian profile.
While the Autocorrelation Function (ACF) exhibits a negative peak at lag $k = 1$, induced by the residual term $\tzt - \g(\tz_{\prev})$ in~\eqref{eq:solution_input_reconstruction}, higher-order lags fall within the 95\% confidence bounds.
\subsection{Online Conditioning}
\label{subsect:online_conditioning}
In order to update online the proxy $\htpolicy_{\prt}$ of the world policy, we leverage the Online Variational Conditioning (OVC) algorithm, originally introduced in \cite{maddox2021conditioning} in the statistics literature.
We summarize the main steps of OVC in Algorithm~\ref{algo:ovc} for a single SVGP, and drop the superscript $i$ from kernel hyperparameters and variational parameters $\prt^i = \{ \eta^i, \inducingmatrix_t^i, \varmean_t^i, \varcov_t^i\}$ when convenient to lighten the notation.
\begin{algorithm}[htpb]
\caption{OVC -- (for a single SVGP)}
\label{algo:ovc}
\begin{algorithmic}[htpb]
    \Require $\eta, \inducingmatrix_{\prev}, m_{\prev}, S_{\prev}$, new sample $(\tz_{\prev}, \hv_{\prev})$
    \State \textbf{Project into pseudo-data space}
    \State $\pcov_{\prev} = \left(S_{\prev}^{-1} - K^{-1}_{\inducingmatrix_{\prev} \inducingmatrix_{\prev}}\right)^{-1}$\hfill\refstepcounter{equation}(\theequation)\label{eq:pseudo_likelihood_cov}
    \State $\ptarget_{\prev} = m(\inducingmatrix_{\prev}) + \pcov_{\prev} S_{\prev}^{-1} (m_{\prev} - m(\inducingmatrix_{\prev}))$ \hfill\refstepcounter{equation}(\theequation)\label{eq:pseudo_targets}
    \\
    \State \textbf{Compute augmented dataset}
    \State $\Xp = \begin{bmatrix}
        \inducingmatrix_{\prev} \\
        \tz_{\prev}^\top
    \end{bmatrix}, \yp = \begin{bmatrix}
        \ptarget_{\prev} \\
        \hv_{\prev}
    \end{bmatrix}, \Sigmap = \begin{bmatrix}
        \pcov_{\prev} & 0 \\
        0 & \vary
    \end{bmatrix}$\hfill\refstepcounter{equation}(\theequation)\label{eq:augmented_dataset}
    \\
    \State \textbf{Update hyperparameters}
    \State $\inducingmatrix_t = $ \textsc{pivchol}$(\Sigmap^{-1/2} K_{\Xp\Xp} \Sigmap^{-1/2})$
    \State $m_t\!=\!m(\inducingmatrix_t) + K_{\Xp \inducingmatrix_t}^\top\!\left(K_{\Xp}\!+\! \Sigmap\right)\inv \!\! (\yp - m(\Xp))$\hfill\refstepcounter{equation}(\theequation)\label{eq:variational_mean_update}
    \State $S_t = K_{\inducingmatrix_t \inducingmatrix_t} - K_{\Xp \inducingmatrix_t}^\top \left(K_{\Xp} + \Sigmap\right)^{-1}K_{\Xp \inducingmatrix_t}$\hfill\refstepcounter{equation}(\theequation)\label{eq:variational_cov_update}
\end{algorithmic}
\end{algorithm}
To provide an intuitive derivation of OVC, we follow the projection view outlined in~\cite{maddox2021conditioning}.
Assume that at time $t$ we start with variational moments $\varmean_{\prev}, \varcov_{\prev}$ and inducing locations $\inducingmatrix_{\prev}$, and want to update these quantities according to a newly observed sample $(\tz_{\prev}, \hv_{\prev})$.
In principle, we could compute updated variational moments $\varmean_t, \varcov_t$ relying on closed-form expressions~\eqref{eq:posterior_inducing_mean}-\eqref{eq:posterior_inducing_cov}.
However, this would (i) require storing an accumulation dataset $\{\tz_j, \hv_j\}_{j = 0}^{\prev}$ and (ii) feature unbounded computational complexity, scaling cubically with $t$ due to the matrix inversion step.
To overcome these limitations, OVC only relies on $\inducingmatrix_{\prev}, \varmean_{\prev}, \varcov_{\prev}$ as sufficient statistics to update the SVGP parameters, without recomputing the entire posterior from scratch at each time step.
The key idea is to project, at each time step $t$, the old parameters $\inducingmatrix_{\prev}, \varmean_{\prev}, \varcov_{\prev}$ into a set of pseudo-quantities, encoding all the information acquired by the SVGP up to time $\prev$. 
Specifically, these are represented by the pseudo-targets $\ptarget_{\prev} \in \RR^M$ and pseudo-likelihood covariance matrix $\pcov_{\prev} \in \RR^{M \times M}$, whose expressions directly follow from the analytical posterior over inducing values.
Intuitively, by setting $\designmatrix = \inducingmatrix$ in equations~\eqref{eq:posterior_inducing_mean}-\eqref{eq:posterior_inducing_cov}, and assuming the likelihood model to be Gaussian, we can solve for $\observationvector, \sigma_y^2 I_\numdata$, resulting in~\eqref{eq:pseudo_likelihood_cov}-\eqref{eq:pseudo_targets}.
By construction, these pseudo-quantities represent a projection of $\pr_{\prev}$ to targets $\ptarget_{\prev}$ and likelihood covariance $\pcov_{\prev}$ for which $\varmean_{\prev}, \varcov_{\prev}$ would be optimal, i.e. equal to $m^\star, S^\star$.
\begin{remark}
    In the limit case where we allocate one inducing point per data sample, equations~\eqref{eq:pseudo_likelihood_cov}-\eqref{eq:pseudo_targets} would exactly recover the entire streaming dataset, i.e., $\inducingmatrix = [\tz_0, \ldots, \tz_{\prev}]^\top$ and $\ptarget_{\prev} = [\hv_0, \ldots, \hv_{\prev}]^\top$.
    \oprocend
\end{remark}
\noindent With these quantities at hand, we can build an augmented dataset~\eqref{eq:augmented_dataset} that accounts for previous data through pseudo-quantities $\inducingmatrix_{\prev}, \ptarget_{\prev}, \pcov_{\prev}$ and the newly acquired sample $(\tz_{\prev}, \hv_{\prev})$.
We underline that, since pseudo-targets $\ptarget_{\prev}$ and reconstructed inputs $\hv_{\prev}$ are governed by different likelihood models, the resulting augmented likelihood covariance $\Sigmap$ is generally heteroscedastic.
Thereby, we leverage the inducing point selection strategy proposed in \cite{maddox2021conditioning} for heteroscedastic noise models, and select new inducing locations $\inducingmatrix_t$ as the first $\numinducing$ pivots of the pivoted Cholesky factorization (\textsc{pivchol} in Algorithm~\ref{algo:ovc}) of the matrix $\Sigmap^{-1/2} K_{\Xp\Xp} \Sigmap^{-1/2}$, where $\Sigmap^{-1/2}$ denotes the inverse of the Cholesky decomposition of $\Sigmap$.
This procedure selects, in a greedy fashion, the $\numinducing$ data points that contribute the most to the posterior variance, as the $j$-th pivot corresponds to the the index with the largest conditional variance given the previously chosen $j-1$ pivots.
Lastly, updated variational mean $m_t$ and covariance matrix $S_t$ are computed by conditioning an exact GP (cf.~\eqref{eq:gp_posterior_mean}-\eqref{eq:gp_posterior_covariance}), with kernel hyperparameters $\eta$, on the augmented dataset $\Xp, \yp, \Sigmap$~\eqref{eq:augmented_dataset}, yielding the expressions in~\eqref{eq:variational_mean_update}-\eqref{eq:variational_cov_update}.
\begin{remark}
    The update rules~\eqref{eq:variational_mean_update}-\eqref{eq:variational_cov_update} have a fixed computational complexity $O(M^3)$, as the augmented dataset~\eqref{eq:augmented_dataset} always contains $M$ inducing points. \oprocend
\end{remark}
\subsection{Trajectory Rollout}
\label{subsect:rollout}
Updated hyperparameters $\prt$ result in an updated predictive posterior $\htpolicy_{\prt}$ which accounts for all samples acquired using information available up to time $\tidx$.
To compute a forecast of the future world system states $\zpredt$ over the prediction horizon $T$, required for our safe control scheme, we need to integrate the nonlinear dynamics~\eqref{eq:world_system} using $\htpolicy_{\prt}(\zt)$ as a proxy of the world policy $\tpolicy(\zt)$.
However, due to the nonlinearity of $\g$ and $\h$, the predictive distribution $\htpolicy_{\prt}$ cannot be propagated analytically through the world system dynamics~\eqref{eq:world_system}. 
Therefore, we resort to a first-order Gaussian moment propagation scheme.
Specifically, we approximate the state distribution at each prediction step as Gaussian and recursively propagate its mean and covariance through a local linearization of the closed-loop world dynamics, as reported in Algorithm~\ref{algo:rollout}.
\begin{algorithm}
\caption{Multi-step Trajectory Rollout}
\label{algo:rollout}
\begin{algorithmic}[htpb]
    \Require Measured world state $\tz_t$, proxy policy $\htpolicy_{\prt}$
    \State Initialize $\hz_{t\mid t} = \tz_t, \hat{\Sigma}^{\z}_{t \mid t} = \varz I_{\targetdimstate}$
    \For{$\mpcidx = 0, \ldots, T-1$}
        \State Evaluate $\htpolicy_{\prt}(\hz_{{\tidx+\mpcidx|\tidx}}) \sim \cN(\postmean_{\prt}(\hz_{{\tidx+\mpcidx|\tidx}}), \postcov_{\prt}(\hz_{{\tidx+\mpcidx|\tidx}}))$
        \\
        \State \textbf{Moment propagation}
        \State $\hz_{{\tidx+\mpcidx+1|\tidx}} = \g(\hz_{{\tidx+\mpcidx|\tidx}}) + \h(\hz_{{\tidx+\mpcidx|\tidx}})\postmean_{\prt}(\hz_{{\tidx+\mpcidx|\tidx}})$\hfill\refstepcounter{equation}(\theequation)\label{eq:mean_propagation}
        \State $A_\mpcidx = \nabla \g(\hz_{{\tidx+\mpcidx|\tidx}})^\top + \nabla \left[\h(\hz_{{\tidx+\mpcidx|\tidx}})\postmean_{\prt}(\hz_{{\tidx+\mpcidx|\tidx}})\right]^\top$
        \State $B_\mpcidx = \h(\hz_{{\tidx+\mpcidx|\tidx}})$
        \State $\hat{\Sigma}^{\z}_{\tidx+\mpcidx+1|\tidx} = A_\mpcidx \hat{\Sigma}^{\z}_{\tidx+\mpcidx|\tidx} A_\mpcidx^\top + B_\mpcidx \Sigma_{\prt}(\hz_{{\tidx+\mpcidx|\tidx}}) B_\mpcidx^\top$\hfill\refstepcounter{equation}(\theequation)\label{eq:covariance_propagation}
    \EndFor
    
\end{algorithmic}
\end{algorithm}

\noindent Our first-order propagation scheme is analogous to an Extended Kalman Filter (EKF) which performs prediction steps only.
At each time step $\tau$ in the prediction horizon, an estimate of the future world system state $\hz_{t+\tau+1|t}$ is computed by propagating the posterior mean $\postmean_{\prt}$ evaluated at $\hz_{t+\tau|t}$ through the nonlinear dynamics~\eqref{eq:world_system}. 
The corresponding predictive covariance $\hat{\Sigma}^{\z}_{t+\tau+1|t}$ is instead evaluated on the linearization of the world system dynamics $\g,\h$ around the current state estimate $\hz_{t+\tau|t}$, yielding the propagation expressions~\eqref{eq:mean_propagation}-\eqref{eq:covariance_propagation}.
We remark that $\hat{\Sigma}^{\z}_{\tidx+\mpcidx+1|\tidx}$ accounts for both the predictive uncertainty of the SVGP posterior and the effect of the (nonlinear) world dynamics on the propagation of uncertainty, captured by the linearization matrices $A_\mpcidx \in \RR^{\dimz \times \dimz}$ and $B_\mpcidx~\in~\RR^{\dimz \times \dimv}$.
Following the product rule for differentiation, $A_\mpcidx$ and $B_\mpcidx$ can be expressed as:
\begin{align*}
    A_\mpcidx &= \nabla \g(\hz_{{\tidx+\mpcidx|\tidx}})^\top + \sum_{i = 1}^{\dimv}\nabla h^i(\hz_{{\tidx+\mpcidx|\tidx}})^\top\postmean^i_{\prt}(\hz_{{\tidx+\mpcidx|\tidx}})\\
    &+ \h(\hz_{{\tidx+\mpcidx|\tidx}})\nabla\postmean_{\prt}(\hz_{{\tidx+\mpcidx|\tidx}})^\top, \\[2ex]
    B_\mpcidx &= \h(\hz_{{\tidx+\mpcidx|\tidx}}),
\end{align*}
where $h^i(\hz_{{\tidx+\mpcidx|\tidx}}) \in \RR^{\dimz}, i \in \{1, \ldots, \dimv\}$ are the columns of the control matrix $\h(\hz_{{\tidx+\mpcidx|\tidx}})$ and $\nabla\postmean_{\prt}(\hz_{{\tidx+\mpcidx|\tidx}}) \in \RR^{\dimz \times \dimv}$ is the gradient of the posterior mean, which can be computed analytically for most commonly used kernel functions, including the RBF we use in this paper.
\subsection{Safe Predictive Control}
\label{subsect:mpc}
\begin{figure}[t]
    \centering
    \includegraphics[scale=1.1]{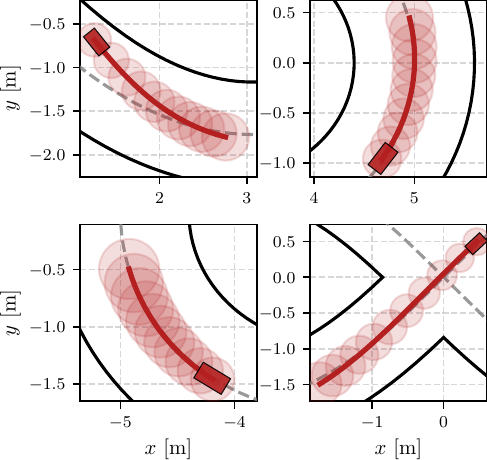}
    \caption{Illustrative example of the world trajectory predictions $\zpredt$ (red) generated by \strategy{}. The red shaded circles (downsampled for clarity) depict the unsafe region defined by collision avoidance constraints~\eqref{eq:collision_avoidance} for $n_\sigma\!=\! 3,T\!=\!30$, while the red rectangle is the current pose of the world robot.}
    \label{fig:rollout_prediction}
\end{figure}

In this section, we derive an implementable, safe MPC scheme, based on the ideal formulation in~\eqref{eq:ego_controller}.
The proposed controller incorporates the predicted world system trajectory $\zpredt$ and its associated predictive uncertainty $\Sigmazpredt$, obtained at each time step $t$ using Algorithm~\ref{algo:rollout}.
First, by denoting the reference state and input trajectories as $\boldsymbol{\x}^{\text{ref}}_t \coloneq \text{col}(\x^{\text{ref}}_{t\mid t}, \ldots, \x^{\text{ref}}_{t+T\mid t})$ and $\boldsymbol{\u}^{\text{ref}}_t \coloneq \text{col}(\u^{\text{ref}}_{t\mid t}, \ldots, \u^{\text{ref}}_{t+T-1\mid t})$, where $\x^{\text{ref}}_{t+\tau \mid t} \in \RR^{\dimx}$ and $\u^{\text{ref}}_{t+\tau \mid t} \in \RR^{\dimu}$, we specialize the stage and terminal cost functions~\eqref{eq:cost} as follows:
\begin{subequations}
\begin{align}
    \ell_\tau &= \|\xtaut - \x^{\text{ref}}_{t+\tau|t}\|^2_Q + \|\utaut - \u^{\text{ref}}_{t+\tau|t}\|^2_R \notag \\
    & \hspace{0.24ex}+\|\utaut - \u_{t+\tau-1 \mid t}\|^2_P + \mathrm{w} \rho_{t+\tau \mid t}^2 \label{eq:stage_cost_specific}\\
    \ell_T &= \|\xTt - \x^{\text{ref}}_{t+\predhorizon|t}\|^2_{Q_T} + \mathrm{w} \rho_{t+\predhorizon \mid t}^2. \label{eq:terminal_cost_specific}
\end{align}
\end{subequations}
where $\rho_{t+\tau \mid t} \in \RR_+$ are slack variables, introduced to enhance the feasibility of the MPC optimization problem.
We set $u_{\prev\mid t} = u_{\prev}$ in~\eqref{eq:stage_cost_specific} and tune the cost parameters $Q, Q_T \succeq 0, R \succ 0, P \succeq 0$ and $\mathrm{w} \in \RR_+$ to achieve a desired trade-off between tracking performance and smoothness of the control input. 
To enforce collision avoidance with the world robot, we represent the physical dimensions of both robots as circles of radius $\radius \in \RR_+$, and inflate the baseline safety distance $2\radius$ by an uncertainty-dependent term.
Specifically, let $p^\x_{t+\tau\mid t} \in \RR^2$ denote the position of the ego robot at time $t+\tau$ in the prediction horizon, and $\hat{p}^\z_{t+\tau\mid t} \in \RR^2$ the position-related component of $\hz_{t+\tau\mid t}$.
We specialize constraints~\eqref{eq:ca_constraints} as:
\begin{subequations}\label{eq:collision_avoidance}
\begin{align}
    \|p^\x_{t+1\mid t}\!-\!\hat{p}^\z_{t+1\mid t}\|^2 &\!\geq\! \left(2r\!+\!n_\sigma \sqrt{\text{Tr}(\hat{\Sigma}^p_{t+1\mid t})}\right)^2, \label{eq:ca_constraints_hard}\\
    \|p^\x_{t+\tau\mid t}\!-\!\hat{p}^\z_{t+\tau\mid t}\|^2 &\!\geq\! \left(2r\!+\!n_\sigma \sqrt{\text{Tr}(\hat{\Sigma}^p_{t+\tau\mid t})}\right)^2\!\!-\!\rho_{t+\tau \mid t} \label{eq:ca_constraints_soft}
\end{align}
\end{subequations}
where $\hat{\Sigma}^p_{t+\tau\mid t} \in \RR^{2 \times 2}$ denotes the position-related block of the covariance matrix $\hat{\Sigma}^{\z}_{t+\tau \mid t}$.
In detail, we enforce the hard constraint~\eqref{eq:ca_constraints_hard} at the first prediction step only, in order to guarantee immediate, strict safety.
For the remainder of the prediction horizon ($\tau = 2, \ldots, \predhorizon$), we relax collision avoidance constraints~\eqref{eq:ca_constraints_soft} through the slack variables $\rho_{t+\tau \mid t}$.
This choice mitigates infeasibilities caused by abrupt changes in the predicted world trajectory $\zpredt$, and allows the MPC to trade temporary safety-margin violations against tracking performance when necessary.
Our formulation~\eqref{eq:ca_constraints_hard}-\eqref{eq:ca_constraints_soft} thus creates an unsafe envelope of circles around $\zpredt$, parameterized by a user-defined safety parameter $n_\sigma \in \RR$, which governs the confidence level of the collision avoidance constraints.
A graphical representation of $\zpredt$ and the unsafe region around it is shown in Fig.~\ref{fig:rollout_prediction} for different time steps, with $n_\sigma = 3$ and $\predhorizon = 30$.
\noindent To summarize, at each time step $t$, our uncertainty-aware MPC control law reads as: 
\begin{equation}
\begin{aligned}\label{eq:ego_controller_specific}
        \min_{\substack{\xtt, \ldots, \xTt \\ \utt, \ldots, \uTt \\ \rho_{t\mid t}, \ldots, \rho_{t+\predhorizon \mid t}}} &\sum_{\mpcidx = 0}^{\predhorizon - 1} \eqref{eq:stage_cost_specific} + \eqref{eq:terminal_cost_specific}\\
        \text{s.t.} \qquad \qquad & \hspace{-7ex} \eqref{eq:dynamics_constraints},\eqref{eq:initcond_constraints},\eqref{eq:input_constraints}, \\
        & \hspace{-7ex} \eqref{eq:ca_constraints_hard}, \\
        & \hspace{-7ex} \eqref{eq:ca_constraints_soft}, \quad \qquad\qquad \, \tau = 2, \ldots, \predhorizon, \\
        & \hspace{-7ex} \rho_{t+\tau \mid t} \geq 0, \quad \qquad\tau = 2, \ldots, \predhorizon, 
\end{aligned}
\end{equation}
\section{Virtual Experiments}
\label{sect:virtual_experiments}
We corroborate the proposed \strategy{} strategy through virtual experiments in an autonomous driving scenario, where the ego and world robots interact in crossing and overtaking maneuvers.
Simulations are performed in Webots~\cite{michel2004cyberbotics} using \textsc{ChoiRbot} \cite{testa2021choirbot}, a ROS 2 Toolbox for distributed robotics, and executed on an Ubuntu 22.04 workstation, powered by an Intel Core i9-13900F CPU and 32 GB of RAM.
The ROS 2 version used in our experiments is Jazzy Jalisco.

\subsection{Virtual Setup}
\begin{figure}[htpb]
    \centering
    \includegraphics[scale=0.26]{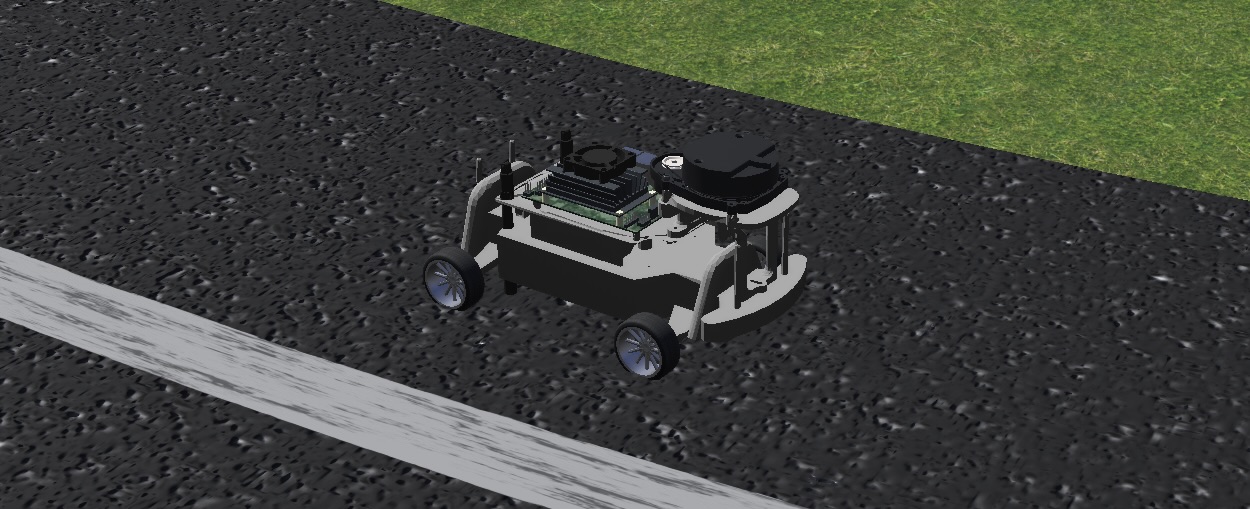}
    \caption{Virtual model of the NVIDIA JetRacer platform in Webots.}
    \label{fig:jetracer}
\end{figure}

In our virtual setup, we select as ego and world robots two NVIDIA JetRacer autonomous vehicles, for which a high-fidelity digital twin, reflecting real-world kinematic and dynamic parameters, is shown in Fig.~\ref{fig:jetracer}. 

\noindent To ensure modularity, each vehicle is managed by an independent ROS 2 control stack, communicating with Webots through a dedicated bridge node.
Specifically, we rely on simplified kinematic models for learning and control purposes (cf. Section~\ref{subsect:models}), while the Webots engine integrates the full, high-fidelity vehicle dynamics.
To emulate a realistic perception pipeline, the ground-truth state of the world vehicle is extracted from Webots and corrupted with zero-mean Gaussian noise ($\sigma_\z = 0.01$), in accordance with the observation model~\eqref{eq:observation_model}.
The resulting noisy measurements $\tzt$ used in Algorithm~\ref{algo:nome_strategia} are fed to the ego vehicle for both offline pretraining and online learning.
\begin{figure*}[t!]
    \centering
    \includegraphics[scale = 0.875]{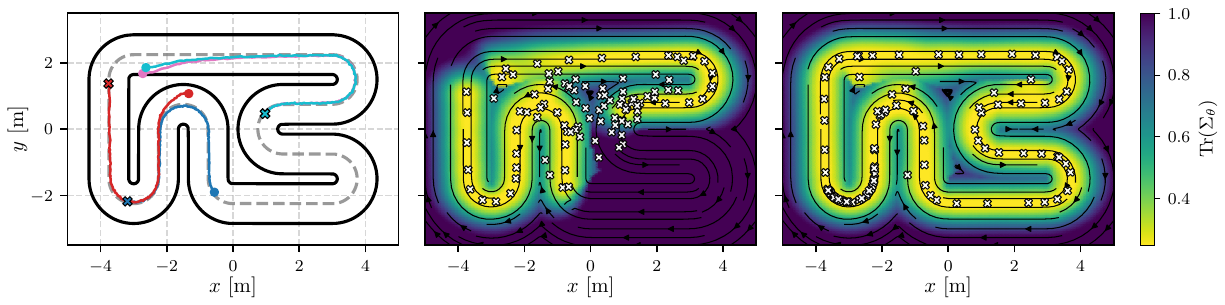}
    \caption{(left) Pretraining world vehicle state trajectories $\mathrm{Z}$ for Section~\ref{subsect:single_experiment}. Estimated vector field $\g(\z) + \h(\z)\htpolicy_{\prt}(\z)$ on the \textsc{ETHZ Mobil} track at $t = 0$ (pretraining, center) and at $t = 2000$ (end of experiment, right). The background colormap quantifies predictive uncertainty, while the white cross-shaped markers denote the inducing locations $\inducingmatrix_t^1$ for the longitudinal velocity SVGP model.}
    \label{fig:pretraining}
\end{figure*}
\subsection{Ego--World Vehicle Models for MPC Design} \label{subsect:models}
We describe the ego vehicle as a kinematic bicycle model with wheelbase $L = 0.17$ m, while the world vehicle is modeled as differential-drive robot, retaining the input-affine structure of the dynamics~\eqref{eq:world_system}.
By employing Forward Euler discretization with sampling time $\Delta t = 50$ ms, the resulting \textsc{ego--world} discrete-time dynamics reads as:
\begin{align}
    \label{eq:ego_dynamics}
    \x_{t+1} &= \xt + \Delta t\begin{bmatrix}
        \ut^1 \cos \varphi^\x_t \\
        \ut^1 \sin \varphi^\x_t \\
        (\ut^1/L) \tan \ut^2
    \end{bmatrix}, \\
    \label{eq:world_dynamics}
    \z_{t+1} &= \zt + \Delta t\begin{bmatrix}
        \cos \varphi^\z_t & 0 \\
        \sin \varphi^\z_t & 0 \\
        0 & 1 
    \end{bmatrix}\vt,
\end{align}
where $\xt = \text{col}(p^\x_t, \varphi^\x_t) \in \RR^3$ and $\zt = \text{col}(p^\z_t, \varphi^\z_t) \in \RR^3$ denote the ego and world vehicles states at time $t$, decomposed into positions $p^\x_t, p^\z_t \in \RR^2$ and orientations $\varphi^\x_t, \varphi^\z_t \in \RR$, while $\ut = [\ut^1, \ut^2]^\top \in \RR^2$ and $\vt = [\vt^1, \vt^2]^\top \in \RR^2$ are the corresponding control inputs.
In all our experiments, the ego and world vehicles are tasked to track the same arc-length parameterized reference trajectory $p^{\text{ref}}(s)$ with assigned speed profiles $\dot{s}(t)$.
For the ego vehicle, following an approach inspired by maneuver regulation schemes~\cite{hauser1995maneuver}, we project the current state $\xt$ at each time step $t$ onto the reference track $p^{\text{ref}}(s)$, yielding:
\begin{equation*}
    \tilde{s} = \tilde{s}(t) = \arg\min_s \|p^{\text{ref}}(s) - p^\x_t\|^2.
\end{equation*}
The reference state and input curves $\boldsymbol{\x}^{\text{ref}}_t, \boldsymbol{\u}^{\text{ref}}_t$ are then generated over the prediction horizon $T$ as:
\begin{gather*}
    s_{t+\tau|t} = \tilde{s} + \tau \u^1_{\text{ref}} \Delta t,  \quad \tau = 0, \ldots, T-1, \\[1ex]
    \x^{\text{ref}}_{t+\tau|t} = \begin{bmatrix}
        p^{\text{ref}}(s_{t+\tau|t}) \\ \psi^{\text{ref}}(s_{t+\tau|t})
    \end{bmatrix}, \: \u^{\text{ref}}_{t+\tau|t} = \begin{bmatrix}
        \u^1_{\text{ref}} \\
        \arctan(L \kappa^{\text{ref}}(s_{t+\tau|t}))
    \end{bmatrix},
\end{gather*} 
where $\u^1_{\text{ref}} \in \RR$ is a desired cruise speed, and the reference yaw $\psi^{\text{ref}}(s)$ and curvature $\kappa^{\text{ref}}(s)$ are related to $p^{\text{ref}}(s)$ by differentiation.

\noindent Regarding the ground-truth world policy, $\tpolicy(\zt)$ is implemented as a path-following controller for unicycle-type robots, adapted from~\cite{carona2008control}.
Specifically, the control law is given by:
\begin{align}
    \label{eq:world_controller}
    \tpolicy(\zt) &= \Delta^{-1} \left(K \tanh(e(s) - \delta) + \cR(\varphi^\z_t)^\top \dot{p}^{\text{ref}}(s)\right), \\
    e(s) &= \cR(\varphi^\z_t)^\top (p^{\text{ref}}(s) - p^\z_t), \notag
\end{align} 
where $\cR(\varphi) \in SO(2)$ is the elementary 2D rotation matrix by an angle $\varphi$, and $\Delta = \text{diag}(1, 0.3)$, $\delta = [0.3, 0]^\top$ and $K = \text{diag}(2, 1)$ are tunable controller parameters.
The resulting linear and angular velocity commands $\vt = \tpolicy(\zt)$ are then mapped into low-level Ackermann steering inputs $[\vt^1, \arctan(L\vt^2/\vt^1)]^\top$ via standard unicycle-to-bicycle kinematic equivalences.

\noindent As for the online learning module, the ego robot is equipped with $\dimv = 2$ independent SVGPs, updated via Algorithm~\ref{algo:ovc}, each comprising $M = 100$ inducing points.
To comply with strict real-time requirements imposed by the sampling time, OVC updates are performed every $5$ time steps, using the most recent batch of data $\{(\tz_{t-5}, \hv_{t-5}), \ldots, (\tz_{t-1}, \hv_{t-1})\}$.
On the other hand, rollouts (Algorithm~\ref{algo:rollout}) are executed at each time step $t$ leveraging the latest available posterior $\htpolicy_{\prt}$.
Nominal MPC parameters~\eqref{eq:ego_controller_specific} are set to $Q = \text{diag}(5, 5, 1), R = \text{diag}(50, 5), P = \text{diag}(1, 50), \mathrm{w} = 1000, Q_T = 2Q$, with actuation limits $\cU = [0.0, 2.5]$ m/s~$\times [-\pi/6, \pi/6]$ rad.
The resulting nonlinear program is formulated in CasADi~\cite{andersson2019casadi} and solved, at each time step $t$, with IPOPT~\cite{wachter2006implementation}.
\subsection{Virtual Experiment on ETHZ Mobil Track} \label{subsect:single_experiment}
We first evaluate the proposed \strategy{} strategy in a realistic racing scenario, selecting as reference trajectory $p^{\text{ref}}(s)$ for both the ego and world vehicles the \textsc{ETHZ Mobil} circuit from~\cite{JainBayesRace2020}.
This experiment is designed to evaluate the adaptation capabilities of \strategy{} when the online observations progressively explore regions of the state space that were not represented in the pretraining dataset $\cD$.
Therefore, we intentionally pretrain the SVGP models over a dataset $\cD$ covering a reduced portion of the overall track, as shown in Fig.~\ref{fig:pretraining} (left).
In detail, $\cD$ consists of $N = 800$ samples collected from $4$ different trajectories of the world vehicle, generated by controlling it using the feedback law~\eqref{eq:world_controller}, and by enforcing a minimum-time speed profile along the centerline via a forward-backward scheme~\cite{kapania2016sequential}.
For each pretraining trajectory, the initial position $p^\z_0$ of the world vehicle is randomly generated along the circuit, while its orientation $\varphi^\z_0$ is tangent to the centerline.
Initial kernel and variational parameters $\pr_0$ are then selected by solving Problem~\eqref{eq:pretraining_elbo} using Adam~\cite{kingma2014adam}, with learning rate empirically set to $\alpha = 0.01$.
At the start of the experiment, we spawn the ego and world vehicle at $p^\x_0 = [-2.5, 2.5]^\top$ m and $p^\z_0 = [2.5, 2.5]^\top$ m respectively, and enforce a constant speed profile $\u^1_{\text{ref}} = 1.7$ m/s along the track for the ego vehicle.
Consequently, $\htpolicy_{\prt}$ is forced to extrapolate, since the world vehicle is outside the region covered by the pretraining data, as shown in Fig.~\ref{fig:pretraining} (center).
\begin{figure*}[t]
    \hspace{0.5ex}
    \includegraphics[scale=1.0]{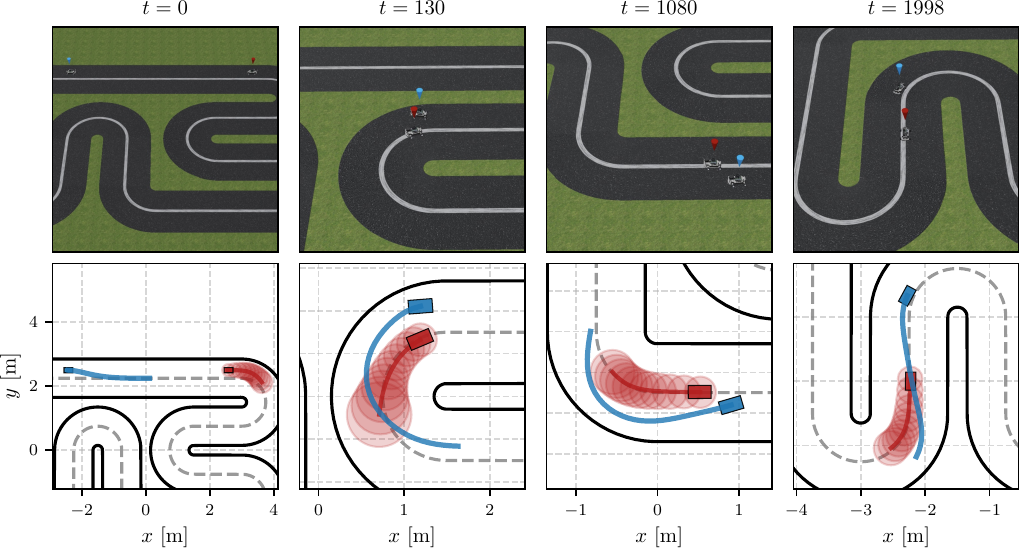}
    \caption{Snapshots at $t \in \{0, 130, 1080,1998\}$ of the virtual experiment performed on the \textsc{ETHZ Mobil} track, with $n_\sigma = 3, T = 30$. Top row: Webots view of the virtual experiment. Bottom row: corresponding top-down view, with the ego planned trajectory (blue), world system predicted trajectory (red) and associated predictive uncertainty (red shaded circles, downsampled for clarity). The reference track $p^{\text{ref}}(s)$ is depicted as a dashed grey line.}
    \label{fig:ethzmobil_snapshots}
\end{figure*}
\begin{figure}[htpb]
    \centering
    \includegraphics[scale=1.0]{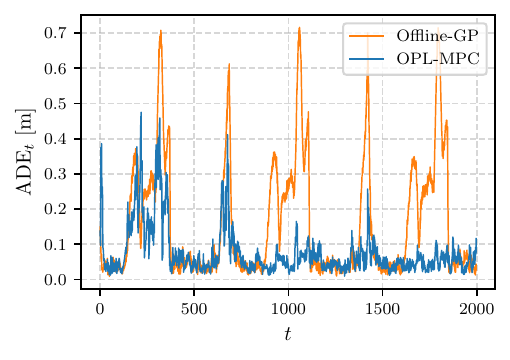}
    \caption{Evolution of the Average Displacement Error (ADE) over $T = 30$ steps across the virtual experiment on the \textsc{ETHZ Mobil} track with (blue) and without (orange) updating the SVGP posterior distribution.}
    \label{fig:lemniscate_ade_evolution}
\end{figure}
\noindent \hspace{-2ex} To quantify prediction accuracy, Fig.~\ref{fig:lemniscate_ade_evolution} shows the evolution of the Average Displacement Error (ADE), computed over the prediction horizon as:
\begin{equation}
    \label{eq:ade}
    \mathrm{ADE}_t = \frac{1}{T} \sum_{\tau = 1}^{T} \|p_{t+\tau}^\z - \hat{p}_{t+\tau \mid t}^\z\|.
\end{equation}
\noindent In particular, we compare the predictions achieved by \strategy{} against an Offline-GP baseline, obtained when $\prt = \pr_0$ are kept fixed to the pretraining values throughout the entire virtual experiment.
Due to unseen regions of the state space, both configurations feature high prediction error during the initial transient phase.
However, after the first lap is completed ($t \approx 700$ steps), the ADE drops and stabilizes to approximately $0.05$~m for \strategy{}, resulting from the variational updates on $\htpolicy_{\prt}$.
This behavior is coherent with the posterior uncertainty shown in Fig.~\ref{fig:pretraining} (right) at the end of the experiment, when the inducing locations $\inducingmatrix$ have been updated to cover the entire circuit, resulting in a significant decrease in the predictive uncertainty.
Conversely, the Offline-GP baseline is unable to assimilate new measurements, resulting in periodic error spikes, reaching above $0.7$ m.
Finally, Fig.~\ref{fig:ethzmobil_snapshots} illustrates relevant snapshots ($t \in \{0, 130, 1080,1998\}$) of the vehicle trajectories in the plane, obtained by interlacing the updated posterior $\htpolicy_{\prt}$ with the proposed uncertainty-aware control stategy~\eqref{eq:ego_controller_specific}.
As expected, \strategy{} is able to guarantee safe overtaking of the world vehicle, thanks to the predictive uncertainty provided by the SVGP models.
\subsection{Monte Carlo Virtual Experiments on Lemniscate Track} \label{subsect:monte_carlo}
\begin{figure*}[t!]
    \centering
    \includegraphics[scale=0.875]{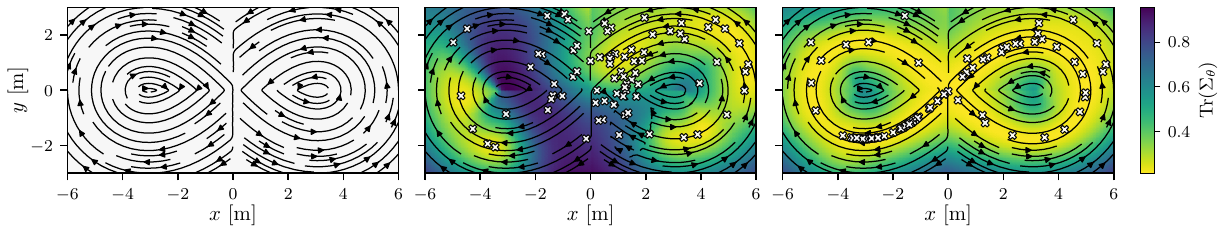}
    \caption{Evolution of the learned world system policy during a representative Monte Carlo trial on the \textsc{Lemniscate} track. (Left) Ground-truth vector field of the closed-loop world dynamics $\g(\z) + \h(\z)\tpolicy(\z)$. (Center) Estimated vector field $\g(\z) + \h(\z)\htpolicy_{\pr_0}(\z)$ after offline pretraining, i.e., at $t = 0$. (Right) Updated vector field $\g(\z) + \h(\z)\htpolicy_{\pr_t}(\z)$ at the end of the simulation, after OVC updates have been performed. The background colormap quantifies predictive uncertainty, while the white cross-shaped markers denote the inducing locations $\inducingmatrix_t^1$ for the longitudinal velocity SVGP model.}
    \label{fig:vector_field_evolution}
\end{figure*}

\begin{figure*}[t]
    \hspace{-1.3ex}
    \includegraphics[scale=1.0]{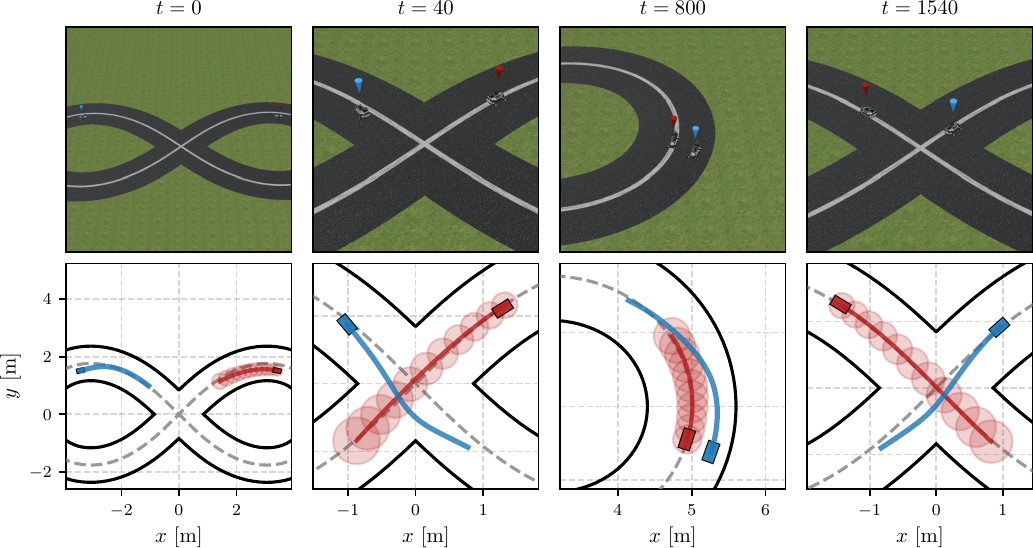}
    \caption{Snapshots at $t \in \{0, 40, 800,1540\}$ of a representative Monte Carlo trial performed on the \textsc{Lemniscate} track, with $n_\sigma = 3, T = 30$. Top row: Webots view of the virtual experiment. Bottom row: corresponding top-down view, with the ego planned trajectory (blue), world system predicted trajectory (red) and associated predictive uncertainty (red shaded circles, downsampled for clarity). The reference track $p^{\text{ref}}(s)$ is depicted as a dashed grey line.}
    \label{fig:lemniscate_snapshots}
\end{figure*}

We assess the statistical robustness of \strategy{} by performing three sets of Monte Carlo virtual experiments on the \textsc{Lemniscate} racing track $p^{\text{ref}}(s)$, described by the following parametric equation:
\begin{equation*}
    p^{\text{ref}}(s) = c + \begin{bmatrix}
        \frac{a\cos (s)}{1 + \sin^2 (s)} \\
        \frac{a\cos (s) \sin (s)}{1 + \sin^2 (s)}
    \end{bmatrix},
\end{equation*}
with scale parameter $a = 5.0$ m and center $c = [0,0]^\top$ m. 
Across all trials, the pretraining dataset $\cD$ is kept fixed and generated as described in Section~\ref{subsect:single_experiment} by randomizing $p^\z_0$ along the track.
Throughout this section, we set $\u^1_{\text{ref}} = 2.0 \mathrm{m/s}$ and consider again the feedback law~\eqref{eq:world_controller} with speed profile generated as in Section~\ref{subsect:single_experiment}.
Figs.~\ref{fig:vector_field_evolution}-\ref{fig:lemniscate_snapshots} illustrate the results of a representative Monte Carlo trial with $n_\sigma = 3$ and $T = 30$.
As discussed in Section~\ref{subsect:offline_pretraining}, the offline pretraining (Fig.~\ref{fig:vector_field_evolution}, center) captures the world system behavior within the available dataset, but yields high predictive uncertainty in unexplored areas of the state space. 
However, as OVC updates are performed online, the inducing locations $\inducingmatrix_t$ are dynamically relocated to cover the newly visited regions. 
This continuously refines the policy proxy $\htpolicy_{\prt}$, which successfully matches the ground-truth vector field (Fig. \ref{fig:vector_field_evolution}, left) and drastically reduces the predictive uncertainty along the \textsc{Lemniscate} track (Fig. \ref{fig:vector_field_evolution}, right).
Fig.~\ref{fig:lemniscate_snapshots} further corroborates the effectiveness of \strategy{} in guaranteeing safe maneuvering of the ego vehicle, in both overtaking ($t = 800$) and crossing scenarios ($t \in \{40, 1540\}$).
\subsubsection*{Varying $n_\sigma$} 
First, we test the effectiveness of \strategy{} in guaranteeing safe maneuvering of the ego robot by varying the safety parameter $n_\sigma \in \{0,1,2,3\}$ with fixed prediction horizon $T = 30$. 
We run $50$ Monte Carlo trials for each configuration, by randomizing the initial positions $p^\x_0, p^\z_0$ of the vehicles along the track.
In Fig.~\ref{fig:mc_varying_nsigma}, we report the Success Rate (SR) and the empirical distribution of the minimum inter-robot distance $d_{\text{min}} \coloneq \min_t\|p^\x_t - p^\z_t\|$ across the trials for different values of $n_\sigma$.
Specifically, when operating without accounting for the predictive uncertainty ($n_\sigma = 0$), our control strategy reduces to an overconfident MPC scheme that assumes $\zpredt$ to be a deterministic forecast, resulting in $\text{SR} = 64\%$ over the $50$ trials.
By increasing $n_\sigma$, the MPC scheme is forced to plan more conservative maneuvers, resulting in a monotone increase in the success rate, reaching $100\%$ for $n_\sigma = 3$.

\subsubsection*{Varying $T$} 
\begin{figure*}[t]
    \centering
    \begin{minipage}{0.48\textwidth}
        \centering
        \includegraphics[width=0.9\linewidth]{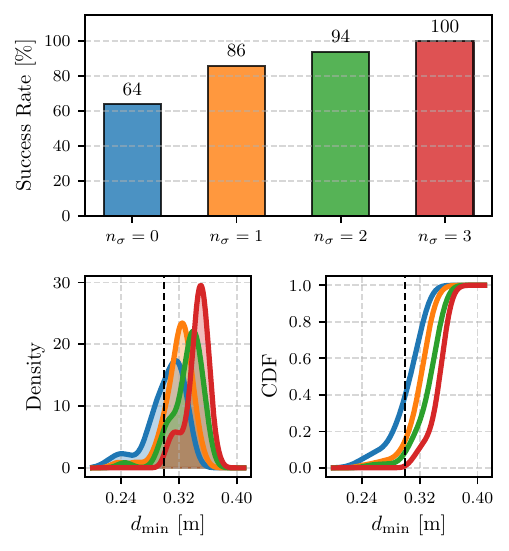}
        \caption{(top) Success Rate of \strategy{} for varying safety parameter $n_\sigma \in \{0,1,2,3\}$. (bottom) Kernel density estimate of $d_{\text{min}}$ and associated Cumulative Distribution Function (CDF) across $50$ Monte Carlo trials. The vertical dashed line represents the safety distance threshold $0.3$~m.} 
        \label{fig:mc_varying_nsigma}
    \end{minipage}\hfill
    \begin{minipage}{0.45\textwidth}
        \centering
        \includegraphics[width=0.9\linewidth]{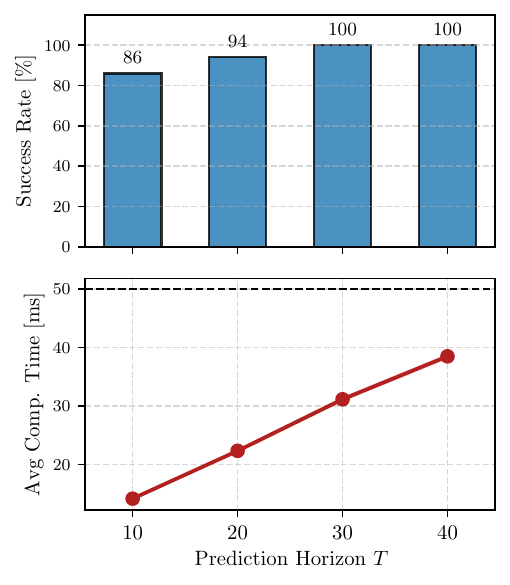}
        \caption{(top) Success Rate of \strategy{} for varying prediction horizon $T \in \{10,20,30,40\}$. (bottom) Average computation time per step required to execute Algorithm~\ref{algo:nome_strategia} for increasing prediction horizon $T$. The horizontal dashed line represents the sampling time $\Delta t = 50$ ms.}
        \label{fig:mc_varying_T}
    \end{minipage}
\end{figure*}

\begin{figure}[htpb]
    \centering
    \includegraphics[scale = 1.0]{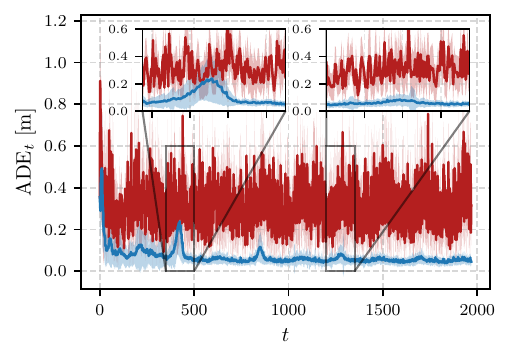}
    \caption{Evolution of the Average Displacement Error (ADE) over $T = 30$ prediction steps for \strategy{} (blue) and CV baseline (red). Results are averaged over $50$ Monte Carlo trials, with solid lines and shaded areas representing the mean and standard deviation, respectively.} 
    \label{fig:comparison_ade}
\end{figure}

Second, we evaluate the real-time feasibility of \strategy{} for increasing prediction horizon values $T\in\{10, 20, 30, 40\}$ with fixed safety parameter $n_\sigma = 3$.
As for the first campaign, we run $50$ Monte Carlo trials for each configuration, by randomizing the initial positions $p^\x_0, p^\z_0$ of the vehicles along the track.
In Fig.~\ref{fig:mc_varying_T}, we highlight the average computation time per time step required to execute Algorithm~\ref{algo:nome_strategia} for different values of $T$, along with the corresponding Success Rate (SR) across the trials.
We observe that increasing $T$ improves the safety performances of \strategy{}, with the success rate monotonically increasing from $86\%$ ($T = 10$) to $100\%$ ($T = 40$).
Indeed, this is expected as, despite the safety buffer provided by $n_\sigma = 3$, higher $T$ allows the uncertainty-aware MPC scheme~\eqref{eq:ego_controller_specific} to plan evasive maneuvers in advance, which better account for the latent world system behavior.
On the other hand, this comes at an increased computational cost of \strategy{}, increasing almost linearly with $T$, despite remaining well below the sampling time $\Delta t = 50 \mathrm{ms}$, confirming the real-time feasibility of our approach even for long prediction horizons.
\paragraph*{Comparison with Baseline} Lastly, we compare the trajectory predictions achieved by \strategy{} ($n_\sigma = 3, T = 30$) against a Constant Velocity (CV) baseline.
To provide a fair comparison, the CV rollouts are performed at each step $t$ by integrating the world system dynamics~\eqref{eq:world_dynamics} with fixed input $\hv_{t-1}$, reconstructed via~\eqref{eq:input_reconstruction}.
We run $50$ Monte Carlo trials for each baseline method, by randomizing the initial positions $p^\x_0, p^\z_0$ of the vehicles along the track.
Results in terms of Average Displacement Error~\eqref{eq:ade} are reported in Fig.~\ref{fig:comparison_ade}, where we observe that \strategy{} significantly outperforms the baseline in terms of prediction accuracy.
Specifically, despite requiring an initial transient phase to observe the behavior of the world system along the entire track, the online updates of \strategy{} allow the ego robot to quickly adapt to the unknown $\pi(\zt)$, resulting in a stable steady-state ADE of approximately $0.05 \mathrm{m}$, versus approximately $0.4 \mathrm{m}$ for CV.
\section{Real Experiments}
\label{sect:real_experiments}
We validate the proposed \strategy{} in an indoor arena ($9 \times 4$ m) using two NVIDIA JetRacer Pro AI Kit autonomous cars.
Specifically, we consider a scenario similar to those described in Section~\ref{subsect:monte_carlo}, where both vehicles track the centerline of a \textsc{Lemniscate} circuit, with scale parameter set to $a = 3.0$ m.
Position and orientation of the vehicles are tracked at $100$ Hz by a $15$-camera Vicon motion capture system.
The world vehicle is controlled by the feedback law~\eqref{eq:world_controller} with a minimum-time speed profile, while the ego vehicle is governed by the proposed \strategy{} with reference velocity $\u^1_{\text{ref}} = 1.7$ m/s.
Longitudinal speed and steering angle commands are then sent to the vehicles at $20$ Hz over a standard $2.4$ GHz WiFi network, using the ROS 2 publisher-subscriber protocol.
Onboard, each car executes a custom ROS 2 low-level node, responsible for converting the Ackermann commands into PWM signals for the actuators.
The hardware workstation, ROS 2 distribution, and optimization solvers are identical to those detailed in Section~\ref{sect:virtual_experiments}.
In the following experiment, we set the MPC parameters~\eqref{eq:ego_controller_specific} to $Q = \text{diag}(1, 1, 1), R = \text{diag}(1, 1), P = \text{diag}(20, 100), \mathrm{w} = 50, Q_T = \text{diag}(1, 1, 2), T = 30, n_\sigma = 3, \cU = [0.0,2.5]$ m/s $\times [-\pi/6, \pi/6]$ rad.
\begin{figure*}[t!]
    \centering
    \includegraphics[scale = 1.0]{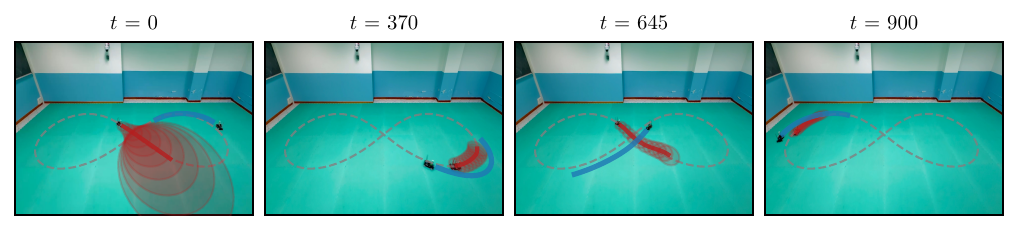}
    \caption{Snapshots at $t \in \{0, 370, 645, 900\}$ of the real experiment, performed on the \textsc{Lemniscate} track, with $n_\sigma = 3, T = 30$. The ego planned trajectory is depicted in blue, while the world vehicle predicted trajectory is shown in red, with the corresponding $n_\sigma = 3$ predictive uncertainty (downsampled for clarity). The reference track $p^{\text{ref}}(s)$ is depicted as a dashed grey line.}
    \label{fig:real_snapshots}
\end{figure*}
\begin{figure}[htpb]
    \centering
    \includegraphics[scale = 1.1]{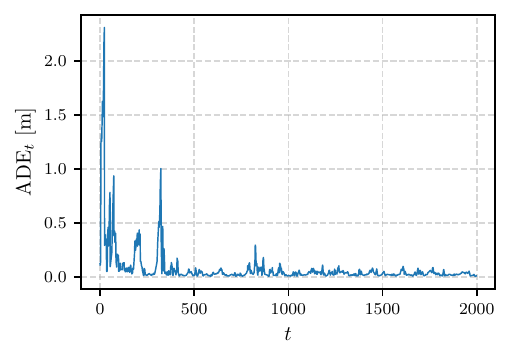}
    \caption{Evolution of the Average Displacement Error (ADE) over $T = 30$ steps across the real experiment.}
    \label{fig:real_ade_evolution}
\end{figure}

In Fig.~\ref{fig:real_ade_evolution}, we illustrate the evolution of the ADE~\eqref{eq:ade} over $T = 30$ across the real experiment, conducted on the \textsc{Lemniscate} circuit in our laboratory.
Coherently with the results observed in the virtual experiments, we observe that the ADE quickly drops after the first lap of the world vehicle ($t \approx 400$), stabilizing at around $0.05$ m.
The snapshots reported in Fig.~\ref{fig:real_snapshots} illustrate the anticipatory avoidance maneuvers enabled by our proposed strategy at $t \in \{0, 370, 645, 900\}$.
Building upon the uncertainty-aware formulation of our control strategy~\eqref{eq:ego_controller_specific}, the ego vehicle deviates in advance from the reference track, avoiding collisions with the world vehicle.

\section{Conclusions}
In this paper, we proposed \strategy{}, a safe learning-based predictive control strategy for \textsc{ego--world} robotic systems, combining an online learning mechanism based on Sparse Variational Gaussian Processes (SVGPs) with a receding-horizon control scheme.
By modeling each component of the unknown world policy as a SVGP, updated via Online Variational Conditioning (OVC) on streaming data, we inferred a posterior distribution over the unknown world policy, which was then propagated through the nonlinear world dynamics using an approximate moment propagation scheme.
The resulting predictive distribution was then incorporated into an uncertainty-aware Model Predictive Control (MPC) scheme, enabling safe maneuvering of the ego robot.
The effectiveness of the proposed architecture was demonstrated through Monte Carlo simulations in ROS 2, and validated on real-world autonomous vehicles in a physical indoor arena.

\section*{Acknowledgments}
\noindent The authors would like to thank Dr. Valerio Digani (SACMI S.C.) for the insightful discussions on the industrial application scenarios that motivated the problem setting, and Andrea Drudi for the support in preparing the real-world experiments.
\bibliographystyle{IEEEtran}
\bibliography{bibfile}

\end{document}